\documentclass[authoryear,3p, preprint,review,11pt]{elsarticle}

\usepackage[T1]{fontenc}
\usepackage{mathptmx}
\usepackage{verbatim}

\usepackage{enumitem}
\usepackage{multirow}
\usepackage{rotating}
\usepackage{epstopdf}
\usepackage{graphicx}
\usepackage{subfigure}
\usepackage{times}

\usepackage{makeidx}
\usepackage{multirow}
\usepackage{multicol}
\usepackage{graphicx}
\usepackage{epstopdf}
\usepackage{ulem}
\usepackage{epstopdf}

\usepackage{array}
\usepackage{amsmath,amssymb,amsfonts}
\usepackage{algorithmic}
\usepackage{algorithm}
\usepackage{graphicx}
\usepackage{textcomp}

\usepackage{epsfig}
\usepackage{makecell}
\usepackage{longtable,booktabs}
\usepackage{color}
\usepackage{longtable}
\usepackage{graphicx}
\usepackage{geometry}

\usepackage{longtable}
\usepackage{multirow}
\usepackage{array}
\usepackage{geometry} 
\usepackage{longtable} 
\usepackage{multirow} 
\usepackage{makecell} 
\usepackage{array} 
\usepackage{ragged2e} 
\usepackage{caption}

\usepackage{natbib}
\usepackage{caption}
\usepackage[colorlinks,
linkcolor=black,       
anchorcolor=black,  
citecolor=black,        
]{hyperref}

\journal{Preprint}

\begin{document}
	\begin{frontmatter}
		
		
		
		\title{Accurate and Robust Generative Approach for Overcoming Data Sparsity and Imbalance in Landslide Modeling with A Tabular Foundation Model}
		
		\author[label1]{Kaixuan Shao}
		\author[label1]{Gang Mei\corref{cor1}}
		\ead{gang.mei@cugb.edu.cn}
		\author[label1]{Yinghan Wu}
        \author[label1]{Nengxiong Xu}
        \author[label1,label2]{Jianbing Peng}
        \cortext[cor1]{Corresponding author}
		
		\address[label1]{School of Engineering and Technology, China University of Geosciences (Beijing), 100083, Beijing, China}
		\address[label2]{School of Geological Engineering and Geomatics, Chang'an University, 710064, Xi'an, China}

	\begin{abstract}
		
		Landslide investigation relies on sufficient and well-balanced observational data influenced by geological, hydrological, and anthropogenic factors. Available landslide inventories are often sparse and imbalanced, which limits understanding of triggering conditions and failure mechanisms. Data generation provides an effective approach to help capture feature dependencies from limited landslide observations. However, existing generation approaches for landslides often struggle to capture complex relationships among features and lack robustness across multiple scenarios and interacting factors. Here, we propose an accurate and robust approach for generating multi-feature landslide datasets by utilizing a tabular foundation model. By leveraging the capacity to learn from limited observations, the proposed approach effectively preserves the multivariate dependencies and statistical characteristics inherent in landslide occurrences. Comparative experiments on 20 landslide inventories demonstrate that the generated datasets closely align with observed distributions, maintain realistic feature dependencies, and exhibit robustness across different environmental contexts. This work provides an effective approach to overcome data sparsity and imbalance and strengthens landslide susceptibility modeling and risk assessment under limited observations.

	\end{abstract}
		

	\end{frontmatter}
	\newpage



\section{Introduction}
\label{sec1}

Landslides are a major class of geohazards that impose widespread social and economic costs through loss of life, damage to infrastructure, and disruption of transport \citep{Glade2012}. Occurring across diverse climates and terrains, landslides often trigger cascading impacts such as river damming, flooding, and long-term landscape destabilization \citep{Guzzetti2012}. Intensifying human pressures and shifting climate regimes are raising exposure and vulnerability worldwide, increasing the demand for robust and transferable methods. Considering that landslides arise from many interacting controls and that each event is unique, naturally available observations and samples are sparse and incomplete even as the overall number of events increases \citep{Steger2016}. Understanding how combinations of geological structure and hydrological drivers produce failure is essential, but the rarity and uneven distribution of observed events make direct discovery of these conditional relationships difficult from field data alone \citep{Hussin2016}. Data limitations for geometric and morphological features are equally important because these features directly control deformation patterns and failure mechanisms \citep{Taylor2018}.

To produce reliable assessments and inform effective mitigation, the observational basis must be both rich and representative. High-quality and multi-feature records enable coverage of diverse landslide failure mechanisms, provide sufficient statistical power to estimate conditional probabilities, and support the transfer of models across different environmental contexts \citep{Sameen2020, Lee2018}. Comprehensive inventories that capture variability in terrain, geology, climate, and human influence are critical for identifying rare but consequential combinations of conditions and for constraining uncertainty \citep{Du2020}. Spatial and temporal balance in sampling is also important to ensure that landslide modeling is not dominated by well monitored, easily accessible, or extreme events. \citep{Ado2022}

However, landslide datasets are typically sparse and strongly imbalanced relative to the vast background of stable terrain \citep{He2009, Galar2012}. Although landslides are widespread in some regions, they remain a small fraction of the overall environmental state space, which creates a pronounced imbalance between landslide and non-landslide samples \citep{Wang2019}. This inherent sparsity complicates data collection and analysis and makes it difficult to capture the full range of heterogeneous landslide behaviors and failure modes, particularly those associated with infrequent conditions or limited monitoring coverage \citep{Zhu2019}.

The sparsity and imbalance of landslide observations limit understanding of triggering conditions and failure mechanisms \citep{He2009}. Existing inventories are often biased toward large or visually prominent failures and toward locations that are easily accessible \citep{Guzzetti2012}. Many records lack the multi-feature detail required for deeper mechanistic analysis \citep{Taylor2018,Micheletti2014}. Remote or difficult to monitor regions such as high-altitude cold environments often lack complete records \citep{Huang2020}. Areas with limited resources or monitoring infrastructure also remain poorly documented \citep{Zhong2020}. These limitations restrict robust characterization of landslide deformation and failure processes \citep{Petschko2014}. Spatial and temporal balance in sampling is also important to ensure that landslide modeling is not dominated by well monitored, easily accessible, or extreme events \citep{Hong2007}. Moreover, data diversity is the foundation of foundation models for earth science. The limited volume and diversity constrain progress in foundation models for landslide science because such models typically rely on large and heterogeneous observational datasets \citep{Bodnar2025, Zhu20262, Ma2021}. These challenges motivate the development of general and reproducible generation methods that produce multi-feature synthetic records conditioned on local environmental context and guided by properties of landslides \citep{AlNajjar2021, Yao2022, Fang2021}. The general and robust methods are important for extracting complex dependencies from limited observations and overcoming data sparsity and imbalance in landslide.

Many studies address landslide data sparsity and class imbalance by working at both the data level and the algorithm level. Statistical and sampling driven methods remain widely used to expand or rebalance landslide inventories and to explore sensitivity to assumed parameter ranges \citep{Breiman1996, Freund1995, Polikar2012, Nanni2015}. Classic Monte Carlo and parametric simulation are applied to generate synthetic attribute values from fitted marginal distributions, while a variety of sampling schemes act directly on observed points to adjust class ratios. Common strategies include random undersampling and informed undersampling of majority class points, random oversampling of minority points, nearest neighbor or cluster based oversampling, and synthetic interpolation methods such as SMOTE and ADASYN \citep{He2009, Chawla2010}. Ensemble resampling techniques such as EasyEnsemble and BalanceCascade have been used in watershed and regional case studies to produce multiple rebalanced training sets \citep{Gupta2022}. Absence selection and spatial sampling design receive particular attention in landslide work, with approaches that include randomly distributed circular absences, similarity sampling, and two-level random schemes intended to control spatial bias when selecting non-landslide examples \citep{Conoscenti2016, Zhu2019}. Several studies combine these data-level maneuvers with adaptive or cluster-aware sampling to better preserve local feature relationships and to create training sets that reflect chosen mapping or modeling objectives \citep{Fang2021, Yao2022}.

Deep learning and generative modeling approaches have been explored to synthesize multivariate landslide records or to learn robust representations under imbalance. Generative adversarial networks and variational autoencoders have been implemented to produce synthetic events and to augment scarce samples for downstream classifiers, with conditional variants used to control generation by trigger type or environmental covariates \citep{AlNajjar2021, Azarafza2021}. More recent work experiments with conditional generation schemes and diffusion-inspired architectures to increase fidelity and mode coverage in synthesized samples. Complementary algorithmic strategies operate at the model level, including cost-sensitive training, class weighting, positive unlabeled learning formulations, and transfer learning from related geoscience tasks, and several studies pursue hybrid pipelines that combine synthetic data generation with ensemble classifiers or domain-aware sampling to improve stability under limited and biased observations \citep{Song2018, Steger2016}.

Despite these efforts, generating meaningful landslide samples from sparse and imbalanced observations remains challenging \citep{Wang2019}. Oversampling at the data level can increase the number of minority samples but may generate synthetic records that violate physically plausible dependencies among features \citep{AlNajjar2021}. Undersampling removes potentially informative background observations and reduces the environmental diversity required for robust generalization \citep{Liu2009,Gupta2020}. Interpolation-based methods such as SMOTE rely on assumptions of local linearity and therefore perform poorly for multimodal and highly heterogeneous joint feature distributions \citep{Haixiang2017}. Algorithmic adjustments, including class weighting and cost-sensitive loss functions, modify decision boundaries but do not enrich the empirical joint feature space, offering limited support for inference on rare or condition-specific failure modes \citep{Galar2012,Tang2009,Nanni2015}. Deep generative models can produce richer synthetic samples, yet their performance depends strongly on the availability of sufficient and representative training data \citep{Azarafza2021,Nikoobakht2022}. They are also sensitive to model architecture and parameter settings and may produce overly concentrated patterns or unrealistic combinations of features that conflict with known physical constraints \citep{Alkhasawneh2018,Althuwaynee2014}. Sampling protocols and absence-selection strategies can alleviate certain biases, but they are often based on empirical rules, tailored to specific regions, and difficult to apply consistently across areas with different survey completeness \citep{Zhu2019,Fang2021,Yao2022}. In addition, label uncertainty, incomplete attribute records, and spatially uneven coverage of multi-feature observations remain largely unresolved. These limitations restrict the ability to reliably characterize the dependencies among landslide-related features and the environmental conditions associated with failure. Therefore, it is important to develop an accurate and robust approach that can capture the complex dependencies and help overcome landslide data sparsity and imbalance \citep{Goetz2015}.  

Recently, foundation models have advanced rapidly and have been applied to an expanding range of geoscientific problems\citep{Bodnar2025, Zhu20261}. In the tabular domain, foundation models pretrained on vast and diverse collections of tables embed broad distributional priors and strong representation capabilities that transfer effectively to small-data tasks\citep{Hollmann2025}. These models process heterogeneous feature types, including continuous covariates, categorical controls, and ordinal attributes. They also learn complex multivariate dependencies during pretraining, which can be leveraged when local observations are sparse. Compared with conventional deep generative models that typically require dataset-specific training and extensive hyperparameter tuning, tabular foundation models substantially reduce the reliance on large local training sets and exhaustive model selection procedures. Therefore, they are well suited to landslide applications, where limited sample sizes, high demands on generalization, and multi-scenario, multi-feature data structures are common. 

Here we propose a simple and robust approach for generating multi-feature landslide datasets by utilizing a tabular foundation model. The key idea is to capture complex dependencies among terrain morphological and trigger related features from limited landslide observations and combine them with broad pretrained multivariate priors so that generated samples remain statistically consistent preserve inter-feature relationships. The approach is demonstrated on a diverse set of rainfall and earthquake-triggered landslide inventories that vary in sample density, feature composition, and geographic setting. Experimental results show that the generated records align closely with observed data and maintain realistic relationships among landslide features, while remaining robust across different landslide scenarios and avoiding common shortcomings of simple interpolation approaches or generative models that require extensive training. 
 
\section{Motivation and Methods}
\label{sec:2}
\subsection{Motivation and key ideas of the proposed approach}
\label{sec:2.1}

Landslide observations are inherently sparse and heavily imbalanced relative to the vast background of stable terrain. Many important landslide failure patterns remain hidden because field observations are spatially biased and often lack consistent multi-feature detail. This intrinsic sparsity and imbalance limit understanding of how combinations of terrain, substrate, and hydrological drivers jointly produce failure. Statistical expansions tend to treat features independently or impose fixed marginal forms and therefore fail to recover multivariate dependencies. At the same time, deep generative methods can capture rich joint structure but often require large and representative training datasets as well as specific model adjustments. An accurate and robust generative approach that expands multi-feature landslide samples while preserving realistic multivariate structure is therefore needed to address data sparsity and imbalance.

Here, we propose a generative approach to address these challenges. The key idea is to capture complex distributional patterns and inter-feature relationships from limited observations based on foundation model, and to generate distinctive and informative samples that help overcome landslide data sparsity and imbalance. Compared with simple oversampling and interpolation methods that may introduce unrealistic relationships, and with deep generative approaches that require large case specific datasets and extensive tuning, this approach is straightforward to implement and maintains strong accuracy and robustness across different types of landslides. It leverages learned multivariate priors to capture dependencies among features from limited observations, generates diverse and physically plausible samples, and preserves clear distinction from stable slope conditions. Through realistic and process consistent data augmentation, sufficient and well-balanced landslide datasets are produced, supporting improved understanding of failure mechanisms and more reliable hazard assessment and management. 

\subsection{Details of the proposed approach}

In this study, we propose a generative approach to capture complex dependencies from limited landslide observations and to generate multi-feature inventories that help alleviate inherent data sparsity and class imbalance. First, landslide records and associated environmental variables are carefully preprocessed so that spatial data, missing values, outliers, and feature scales are handled in a consistent manner, and the resulting datasets are converted into structured tabular form. Second, a tabular foundation model generates candidate landslide samples by learning the joint structure of the local inventory and producing new rows through sequential conditional sampling. Third, the candidate pool is filtered and integrated using a plausibility-based selection rule, after which the accepted samples are merged under controlled mixing. After these steps, the generated samples remain statistically consistent with the observed inventory and preserve multivariate dependencies among topographic, hydrological, and geometric attributes. Multiple complementary metrics are then used to evaluate and compare the accuracy and robustness of the proposed approach across different landslide scenarios.

\label{sec:2.2}

\begin{figure}[H]
	\centering
	\includegraphics[width=\textwidth]{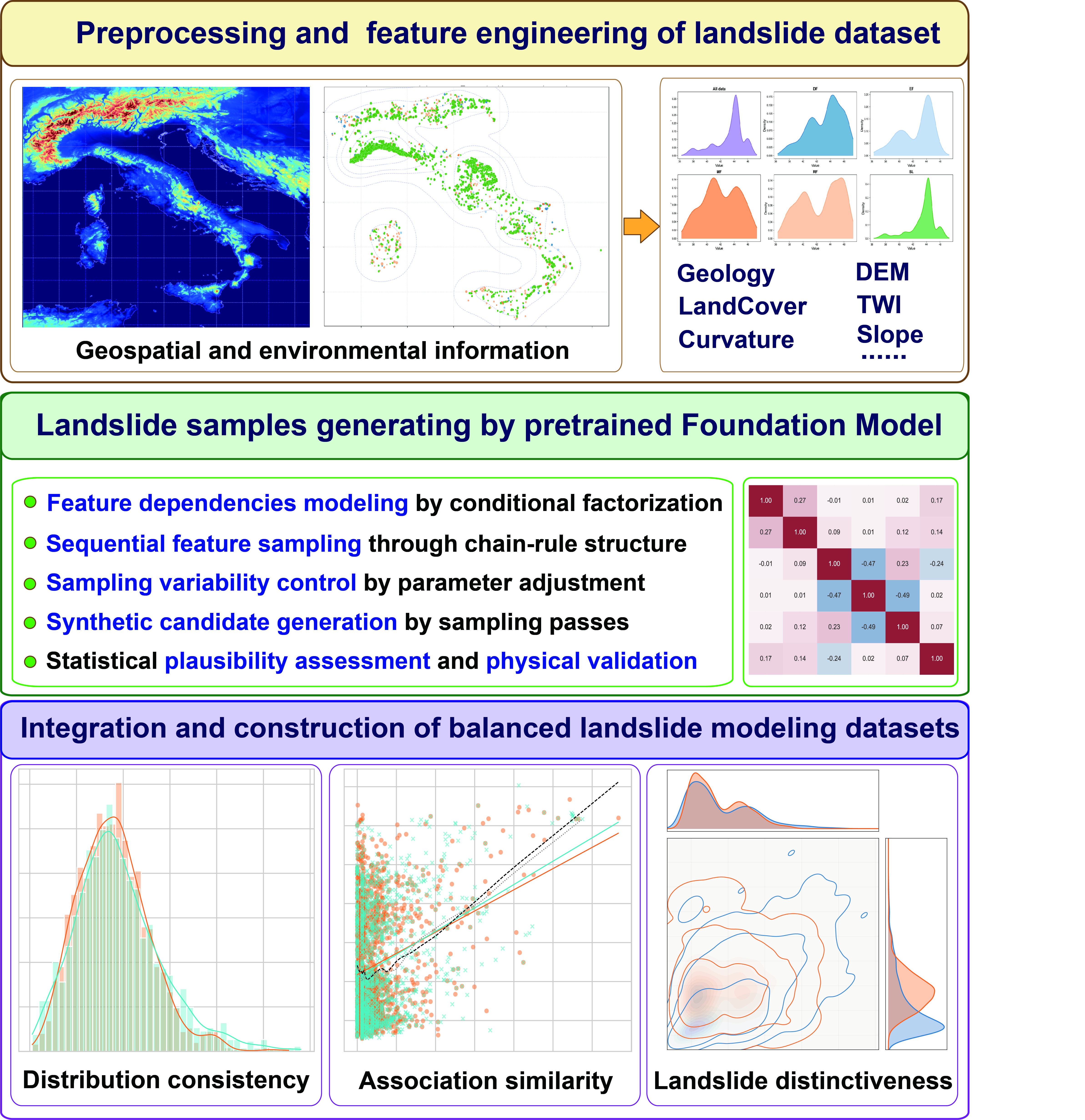}
	\caption{Workflow of the proposed foundation model-based approach for overcoming landslide data sparsity and imbalance}
	\label{fig:1}       
\end{figure}

\subsubsection{Step 1: Preprocessing and extracting landslide features}
\label{sec:2.2.1}

Considering the inherent complexity of landslide processes and the requirements of quantitative analysis, landslide inventories and associated environmental variables require careful preprocessing to ensure internal consistency. Typical steps include assembling records from multiple sources, screening cases for completeness and spatial accuracy, and applying transparent rules to handle missing information. Spatial datasets are unified to a common coordinate system and resolution, and geometric and topological checks are performed to remove duplicates and correct invalid features. Missing values are explicitly marked using indicator variables, while imputation is applied selectively and in a controlled manner to avoid distorting underlying patterns. Simple deterministic methods such as median or k nearest neighbor imputation are used for limited and randomly distributed gaps, whereas more structured missingness is addressed using approaches such as multiple imputation. At the same time, extreme values and outliers are examined and treated to reduce their undue influence while preserving meaningful variability. Continuous attributes that exhibit skewness or heavy tails are transformed to stabilize variance and improve conditional modeling behavior, including logarithmic transformation for positive variables, rank or quantile transformation for heavy tailed distributions, and z score normalization to ensure comparability across features. These preprocessing steps ensure that the resulting datasets are consistent, interpretable, and suitable for capturing multivariate relationships in landslide analysis.

Influenced by different research objectives and data availability, most landslide inventories emphasize distinct aspects beyond basic spatial and temporal attributes, including terrain morphology and triggering conditions. For geomorphological analyses, terrain variables derived from digital elevation models are commonly used, such as slope, curvature, topographic position index, and topographic wetness index, with attention to spatial resolution and window size due to scale dependence. For studies focused on triggering mechanisms, temporal variables such as rainfall and soil moisture are aggregated into antecedent indicators using defined accumulation periods and processing rules. For analyses of landslide geometry and runout behavior, polygon based metrics including area, length, orientation, and shape descriptors are extracted, often using object based approaches when landslide boundaries are clearly defined. Following these steps, datasets with different emphases are systematically transformed through feature extraction, encoding, and standardization into structured tabular formats, where each row represents an individual landslide and columns correspond to numerical and categorical attributes. Such representations support downstream analyses including susceptibility modeling, identification of triggering conditions, and quantitative evaluation of controlling factors. In this study, twenty rainfall and earthquake triggered landslide inventories from different regions and with varying feature compositions are compiled to evaluate the performance of the proposed approach across diverse scenarios.

\subsubsection{Step 2: Generating landslide samples with a tabular foundation model}
\label{sec:2.2.2}

This study uses the unsupervised generation mode of the tabular foundation model to produce candidate landslide samples from the local inventory. The generator is first exposed to the local table so that its internal predictors reflect the empirical structure of the observed data. It then proposes new rows by modeling the joint distribution of the features through conditional factors and by drawing each feature sequentially from the learned conditionals.

For a $d$ dimensional row $x = (x_1,\dots,x_d)$, the joint distribution is represented by the chain rule as

\begin{equation}
p(x_1,\dots,x_d) = \prod_{i=1}^{d} p\bigl(x_i \mid x_1,\dots,x_{i-1}\bigr)
\end{equation}

Under this factorization, each conditional term describes the distribution of one feature given those that have already been sampled. This is the statistical basis for generating coherent rows that preserve multivariate dependencies across continuous and categorical variables. The generator therefore produces each candidate by sequentially sampling the conditional factors in turn.

Sampling diversity is controlled by a temperature parameter $T$. The conditional draw for each feature is written as
\begin{equation}
x_i \sim p_T\bigl(x_i \mid x_1,\dots,x_{i-1}\bigr), \qquad
p_T(\cdot) \propto p(\cdot)^{1/T}
\end{equation}

where larger values of $T$ increase diversity and smaller values concentrate the draw near high probability regions of the learned conditional distribution. Repeating this sequential sampling process yields a candidate pool of size $N$. In addition, the generator computes a permutation averaged plausibility proxy by evaluating the factorized likelihood under multiple random feature orders and averaging the results,
\begin{equation}
\hat{p}(x) = \frac{1}{M} \sum_{m=1}^{M} p_{\text{model}}\bigl(\pi_m(x)\bigr)
\end{equation}

where $\pi_m$ is the $m$th random permutation of feature indices and $M$ is the number of permutations averaged. This score provides a stable ranking signal that is less sensitive to feature order and is therefore suitable for organizing the generated candidates before selection.

The generation process is summarized by three controls that are directly reflected in the TabPFN generation workflow. The first is the requested candidate count $N$, which determines how many synthetic rows are produced. The second is the sampling control $T$, which determines how exploratory the generated samples are. The third is the number of permutations $M$, which determines how stable the plausibility estimate is. Generation can also be focused on selected attributes when only certain columns require augmentation, while the remaining columns continue to provide context for the conditional draws. Every candidate is retained together with its generation context and plausibility score so that the produced rows remain reproducible and traceable.

\subsubsection{Step 3: Selection and Integration of landslide samples}
\label{sec:2.2.3}

The candidate pool produced in the previous step is then filtered and integrated into the study dataset. The selection stage begins from the finite set $\mathcal{C} = {x^{(1)},\dots,x^{(N)}}$ and uses the permutation averaged plausibility score as the main ranking basis. Because the log form is numerically more stable and easier to compare across candidates. A simple acceptance rule is then applied. For a chosen threshold $\tau$, candidates are retained when
\begin{equation}
\log \hat{p}(x) \ge \tau
\end{equation}

This criterion selects the rows that are most consistent with the joint structure learned by the generator while excluding low plausibility proposals. The accepted subset is denoted by $\mathcal{A} \subseteq \mathcal{C}$. After selection, the accepted rows are converted into canonical table form, their feature types are aligned with the original inventory, and their generation metadata are recorded. The metadata include the candidate count $N$, the sampling control $T$, the number of permutations $M$, the conditioning context, and the plausibility score $\log \hat{p}(x)$. This keeps the synthetic rows auditable and allows the generation history of each accepted sample to be tracked.

The accepted samples are then integrated by controlled mixing. Let $\alpha$ denote the fraction of synthetic rows introduced into the final corpus. A smaller $\alpha$ keeps the training set closer to the observed inventory, while a larger $\alpha$ increases the contribution of synthetic samples to underrepresented regions of the feature space. The integration step preserves the original feature encodings and does not alter the underlying feature definitions. Its role is to expand the available landslide sample set in a controlled manner so that the final corpus contains both observed and accepted synthetic examples with transparent provenance.

After these steps, the generated and accepted samples remain statistically consistent with the observed inventory, preserve multivariate dependencies among topographic, hydrological, and geometric attributes, and enlarge the representation of sparse parts of the joint feature space. The generated samples are statistically consistent with observations, preserve multivariate dependencies among topographic, hydrological, and geometric attributes, and expand underrepresented regions of the joint feature space. This process helps alleviate data sparsity and imbalance and improves understanding of landslide failure processes and triggering conditions.

\subsection{Comparison and evaluation}
\label{sec:2.3}

\subsubsection{Baseline methods for comparison}
\label{sec:2.3.1}

There are three approaches are employed as the comparative baselines to evaluate the performance of the proposed generative approach.

(1) Monte Carlo methods

Monte Carlo methods generate synthetic samples by sampling from an explicit probability model for the feature vector. The usual steps are estimate the marginal and conditional distributions of relevant variables, incorporate domain priors and constraints where possible, and then draw samples that reflect those distributions. This approach is statistical in nature and makes uncertainty explicit, which makes it straightforward to encode prior geological knowledge and to control the generated sample statistics.
In practice, the Monte Carlo methods consist of distribution estimation, sampling and postprocessing to enforce categorical constraints and physical plausibility. The method requires specification of the joint or conditional model for the data and verification that the generated samples preserve relevant marginal and conditional statistics.

(2) SMOTE

SMOTE creates new minority class examples by interpolating between nearby observed minority instances in feature space. The core idea is to increase minority class density without simple duplication by synthesizing points along the lines that connect a minority sample to its nearest minority neighbors. This procedure is conceptually simple and computationally light, and it is commonly used as a preprocessing step before training standard machine learning classifiers. The process for generating sample is as follows:
\begin{equation}
\mathbf{x}_{\text{new}} = \mathbf{x}i + \lambda \left( \mathbf{x}{\text{nn}} - \mathbf{x}_i \right)
\end{equation}
In this expression $\mathbf{x}i$ denotes a minority class sample, $\mathbf{x}{\text{nn}}$ denotes one of its $k$ nearest minority neighbors, and $\lambda$ denotes a scalar drawn from the uniform distribution on the interval $[0,1]$. SMOTE assumes that linear interpolation in the chosen feature representation produces plausible minority cases and therefore requires careful examination when variables are mixed or physical constraints are strong.

(3) GAN

As a deep learning approach, GAN learns to produce synthetic samples by training a generator network to mimic the data distribution while a discriminator network learns to distinguish real from generated samples. The training is adversarial and data driven, which allows the model to capture complex nonlinear joint distributions and high order dependencies among features. Conditional GAN extend the framework by incorporating auxiliary information such as class labels or environmental attributes to guide the generation process toward specific subsets of the data, including minority classes or particular geological or climatic conditions. The adversarial learning process is commonly formulated as a minimax optimization problem that defines the objective function governing the interaction between the generator and discriminator as follows
\begin{equation}
\min_{G}\max_{D} V(D,G)
\mathbb{E}{\mathbf{x}\sim p{\text{data}}}\left[\log D(\mathbf{x})\right]+
\mathbb{E}{\mathbf{z}\sim p{z}}\left[\log\left(1 - D\left(G(\mathbf{z})\right)\right)\right]
\end{equation}

In the conditional setting, the generator produces samples of the form $G(z,y)$ while the discriminator evaluates pairs $D(x,y)$, allowing the model to learn distributions conditioned on specific variables. Although GANs are capable of generating realistic and diverse samples, their performance depends strongly on model configuration and the availability of sufficient and representative training data, and they may produce inconsistent or physically implausible feature combinations under limited or complex data conditions.

\subsubsection{Evaluation metrics}
\label{sec:2.3.2}

To evaluate whether the generated landslide samples faithfully reproduce the statistical structure of the original observations, we used a compact set of complementary metrics that capture center, spread, and distributional shape from different angles. The original landslide data are sparse and imbalanced, so a single score is not sufficient for judging quality. We therefore combined measures of location shift, dispersion mismatch, distributional distance, and overall shape similarity. Together, Absolute error, Relative error, Standard deviation difference, Relative standard deviation difference, Bias per SD, Wasserstein distance per SD, KS statistic, and JS divergence provide a balanced assessment of whether the generated samples remain close to the original landslide observations while still preserving realistic variability and joint data behavior.

The first group of metrics focuses on the central tendency of each feature. The Absolute error measures the direct difference between the mean of the original data and the mean of the generated data, and the Relative error expresses the same difference as a fraction of the original mean. These two metrics reflect whether the generated data are systematically shifted upward or downward relative to the original landslide observations. Absolute error is expressed in the original unit of the feature, so it is easy to interpret physically, while Relative error is scale free and therefore suitable for comparing features with very different magnitudes. In this study, these two quantities are especially useful for identifying whether the generator reproduces the typical level of terrain, hydrological, or geomorphic variables without introducing a biased center.

The second group of metrics evaluates the spread of the generated samples. Standard deviation difference measures whether the generated data are too concentrated or too dispersed relative to the original observations, and Relative standard deviation difference scales that difference by the original standard deviation. These metrics complement the mean based scores because a method may match the center of a feature while still failing to preserve its variability. The Bias per SD metric strengthens this idea by measuring mean error in units of the original standard deviation. It is defined as follows

\begin{equation}
\text{Bias per SD} = \frac{\left| \bar{x}{\text{orig}} - \bar{x}{\text{gen}} \right|}{s_{\text{orig}} + \epsilon}
\end{equation}

This normalization is important because it places all features on a common scale and makes the magnitude of the bias easier to interpret across variables with different units. A small value means the generated mean differs only slightly from the natural spread of the original data, whereas a large value indicates that the shift is large relative to the inherent variability of the feature. In the context of landslide sample generation, these spread related measures are important because preserving variability is essential for representing diverse failure conditions rather than producing overly smooth or overly narrow synthetic samples.

We also used distributional distance measures that capture differences beyond the mean and standard deviation. The Wasserstein distance per SD quantifies the average transport cost needed to transform the generated distribution into the original one, after standardizing by the original standard deviation. It is defined from the one dimensional Wasserstein distance as follows

\begin{equation}
W_1(P,Q) = \int_{0}^{1} \left| F_P^{-1}(u) - F_Q^{-1}(u) \right| , du
\end{equation}

\begin{equation}
\text{Wasserstein distance per SD} = \frac{W_1(P,Q)}{s_{\text{orig}} + \epsilon}
\end{equation}

This metric is especially useful because it reflects both location shifts and broader shape differences in a single quantity, while still being comparable across variables after standardization. The KS statistic is another global distributional measure, defined as the maximum distance between the empirical cumulative distribution functions of the original and generated samples. It is given as follows

\begin{equation}
D = \sup_{x} \left| F_{\text{orig}}(x) - F_{\text{gen}}(x) \right|
\end{equation}

Unlike the mean based metrics, the KS statistic is sensitive to mismatches anywhere in the distribution, including differences in tails or multimodal structure. A small KS statistic means the two empirical distributions are close throughout their range, whereas a larger value indicates a more substantial mismatch. In our landslide application, these two distance based metrics are valuable because they reveal whether the generator preserves realistic statistical transport and cumulative structure, not just average values.

Finally, the JS divergence provides a complementary view of similarity from an information theoretic perspective. After estimating the two feature distributions with kernel density estimation on a common grid, the distributions are normalized and compared through the Jensen Shannon divergence. Its conceptual definition is as follows

\begin{equation}
\mathrm{JS}(P,Q) = \frac{1}{2}\mathrm{KL}(P,M) + \frac{1}{2}\mathrm{KL}(Q,M)
\end{equation}

\begin{equation}
M = \frac{1}{2}(P + Q)
\end{equation}

Because JS divergence is symmetric and bounded, it offers a stable measure of how similar the overall density shapes are. A smaller value indicates that the generated data allocate probability mass in a way that is close to the original observations, while a larger value indicates a stronger departure in distribution shape. Together with the Wasserstein distance per SD and KS statistic, JS divergence helps assess whether the generated landslide samples preserve realistic multivariate patterns and not merely marginal averages. In this way, the full metric set provides a structured evaluation of whether the synthetic data remain statistically close to the original landslide observations while still being sufficiently informative for downstream analysis and hazard assessment.

\section{Results}
\label{sec:3}

\subsection{Landslide inventories}
\label{sec:3.1}

Landslides are governed by interacting factors that vary across space and triggering conditions, and no single dataset can fully characterize these interactions. To reflect the intrinsic complexity of landslide processes and to evaluate the generality of the proposed approach, a collection of landslide datasets was compiled spanning multiple regions, triggers, and feature emphases. Rather than focusing on a single well sampled area or a single type of attribute, the datasets were selected to represent different perspectives on landslide occurrence and behavior. Some datasets emphasize triggering mechanisms, others focus on terrain and environmental controls, while others capture geometric and morphological expressions of failure. These heterogeneous datasets provide a meaningful basis for testing whether the generation approach remains robust and transferable across scenarios with different feature structures and sample densities.

The rainfall-induced landslide datasets collectively represent a broad spectrum of hydro-geomorphic settings and observational emphases. Several datasets are designed to characterize how terrain and environmental conditions modulate rainfall-driven instability. These emphasize topographic attributes such as elevation, slope, curvature, hydrological indices, land cover, lithology, and proximity to infrastructure, and are commonly used for susceptibility assessment and spatial prediction. Other datasets place stronger emphasis on rainfall as a trigger, incorporating multi-temporal precipitation indicators that capture short-term intensity and long-term antecedent moisture conditions. A third group focuses on landslide geometry and morphology, providing polygon-based descriptors of size, shape, orientation, and slope context that are closely linked to failure mechanisms and post-failure evolution.

By integrating these complementary perspectives, the rainfall-induced datasets allow investigation of landslides from triggering, conditioning, and manifestation viewpoints. Scientifically, this supports analysis of how rainfall interacts with terrain and material properties to produce different modes of failure. For data generation, it creates a heterogeneous test environment in which synthetic records must respect both environmental dependencies and geometric plausibility. The presence of both small, detailed inventories and large regional compilations further enables evaluation under varying degrees of data sparsity and imbalance, which are important in rainfall-induced landslide modeling.

\subsection{Scenario I: Landslide datasets focused on terrain and geomorphological characteristics}
\label{sec:3.2}

We first evaluated generation fidelity across representative rainfall triggered landslide inventories using a set of terrain and geomorphological factors such as slope curvature wetness related indices distance to infrastructure and other environmental characters. Synthetic outputs from TabPFN, CTGAN, SMOTE and Monte Carlo sampling were compared under a unified preprocessing and evaluation protocol with consistent feature encoding. Kernel density estimates histogram overlays conditional summaries and joint scatterplots were used to examine whether the generated landslide samples reproduce observed feature characteristics and capture complex dependencies among variables.

\begin{figure}[H]
	\centering
	\includegraphics[width=\textwidth]{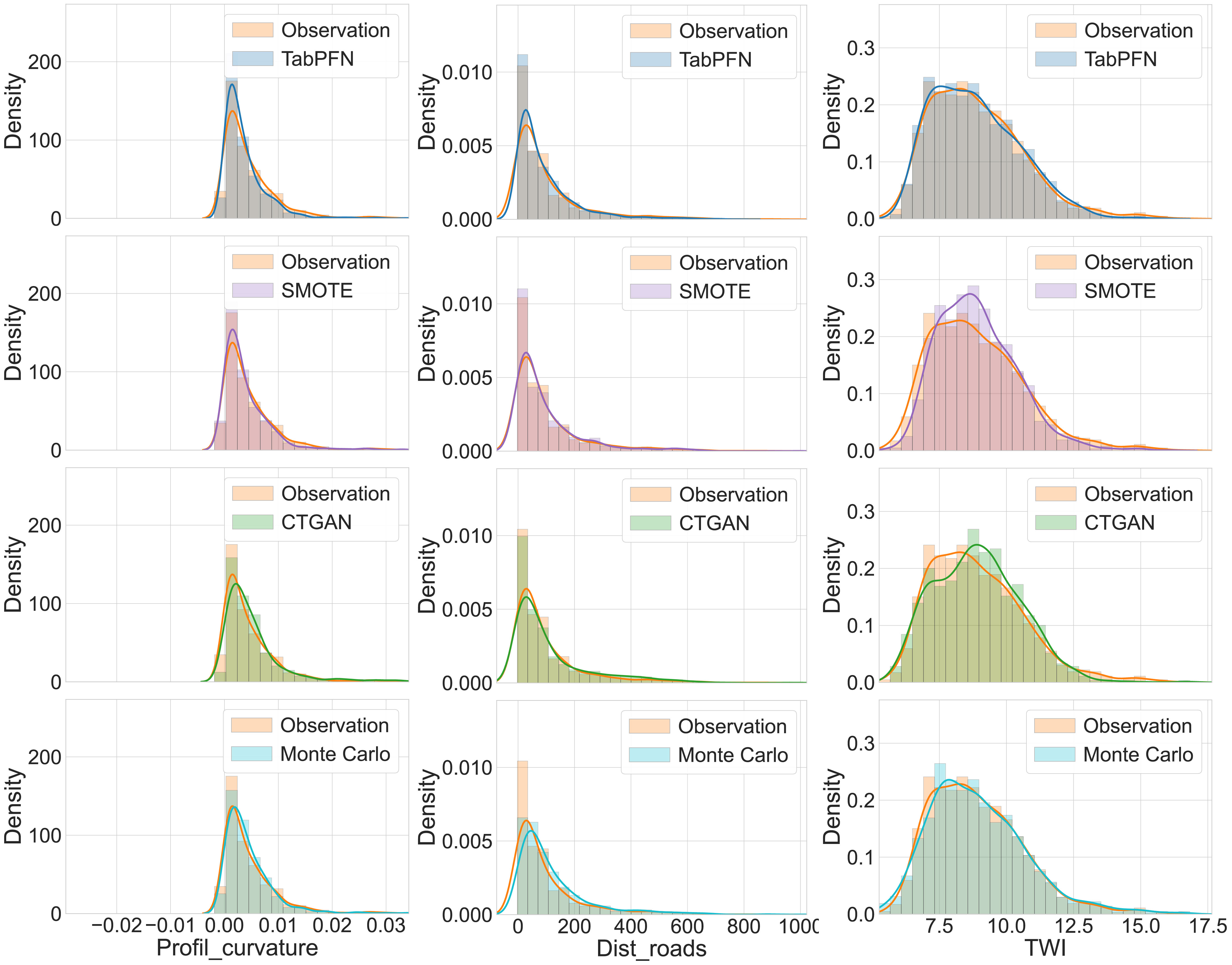}
	\caption{Statistical characteristics of terrain and geomorphological features in rainfall-triggered landslide inventories generated in Italy}
	\label{fig:2}       
\end{figure}

The selected predictors exhibit clearly different empirical patterns in the observed inventory. Profile curvature shows a pronounced peak at zero with a symmetric spread around that peak. Distance to roads displays a heavy tailed distribution with high density near zero and rapid decay at larger distances. Topographic wetness index presents a complex irregular density without a dominant single mode. For features with clear and well structured marginal shapes all four generators reproduce the main univariate characteristics. Profile curvature and distance to roads are captured reasonably well by each method. In the case of distance to roads CTGAN achieves particularly close agreement with the observed tail behavior which indicates that conditional neural sampling can model skewed marginals when sufficient structure is present.

Differences among methods become evident for weakly structured or noisy marginals. For topographic wetness index TabPFN best preserves the subtle and irregular empirical density without introducing peaks that are not observed in the original data. Monte Carlo sampling provides a moderate match but lacks fine structural detail. In contrast CTGAN and SMOTE tend to produce additional modes or overly smoothed peaks that do not occur in the observation. These patterns suggest that interpolation based oversampling and dataset specific adversarial training may impose modal structures that are not supported by the data under complex distributions whereas pretrained multivariate priors allow TabPFN to better respect irregular empirical forms. 

\begin{figure}[H]
	\centering
	\includegraphics[width=\textwidth]{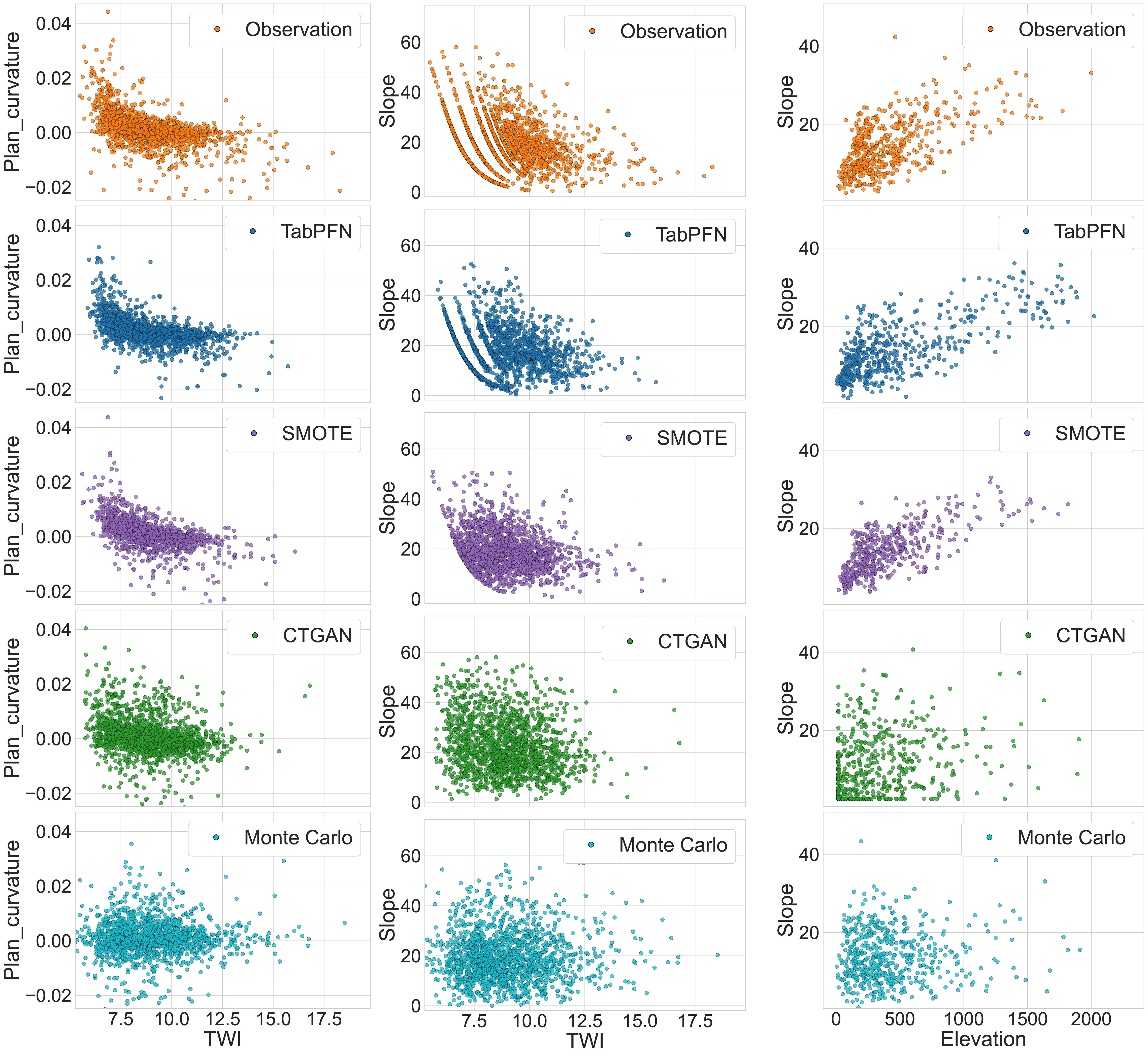}
	\caption{Feature dependence of terrain and geomorphological characteristics in rainfall-triggered landslide inventories generated in Italy}
	\label{fig:3}       
\end{figure}

Beyond univariate fidelity realistic landslide augmentation requires preservation of process relevant inter feature relationships. In the rainfall inventory topographic wetness index co varies systematically with slope and plan curvature. Higher wetness index values correspond on average to lower slope angles and reduced plan curvature which reflects the hydro geomorphic signature of rainfall induced failures in the study region. TabPFN generated samples closely reproduce this conditional trend without additional local retraining. SMOTE reflects the overall direction of association but smooths local variability and attenuates conditional variance due to its interpolative mechanism. CTGAN does not consistently recover the observed joint trend when the conditional pattern is complex and the effective local training signal is limited. Monte Carlo sampling matches marginal distributions yet fails to capture the observed multivariate dependence. These differences imply that methods which only match marginals or which rely on local interpolation may misrepresent process relevant couplings that are critical for mechanistic inference.

In sparse regional inventories such as the Loess Plateau, the azimuth density reveals gaps in the observed record that challenge simple interpolation, TabPFN generates samples that occupy azimuth ranges where observations are scarce or absent, this behavior contrasts with SMOTE which by construction can only interpolate between existing observations and therefore cannot populate genuinely missing azimuth bins, the ability of TabPFN to propose plausible values in underrepresented azimuth ranges suggests that pretrained multivariate priors supply complementary statistical information that helps infer plausible directional variation, this property improves the representativeness of augmented inventories for analyses that are sensitive to failure orientation, and it reduces the risk that downstream models will systematically ignore conditional modes absent from small surveys.

\begin{figure}[H]
	\centering
	\includegraphics[width=\textwidth]{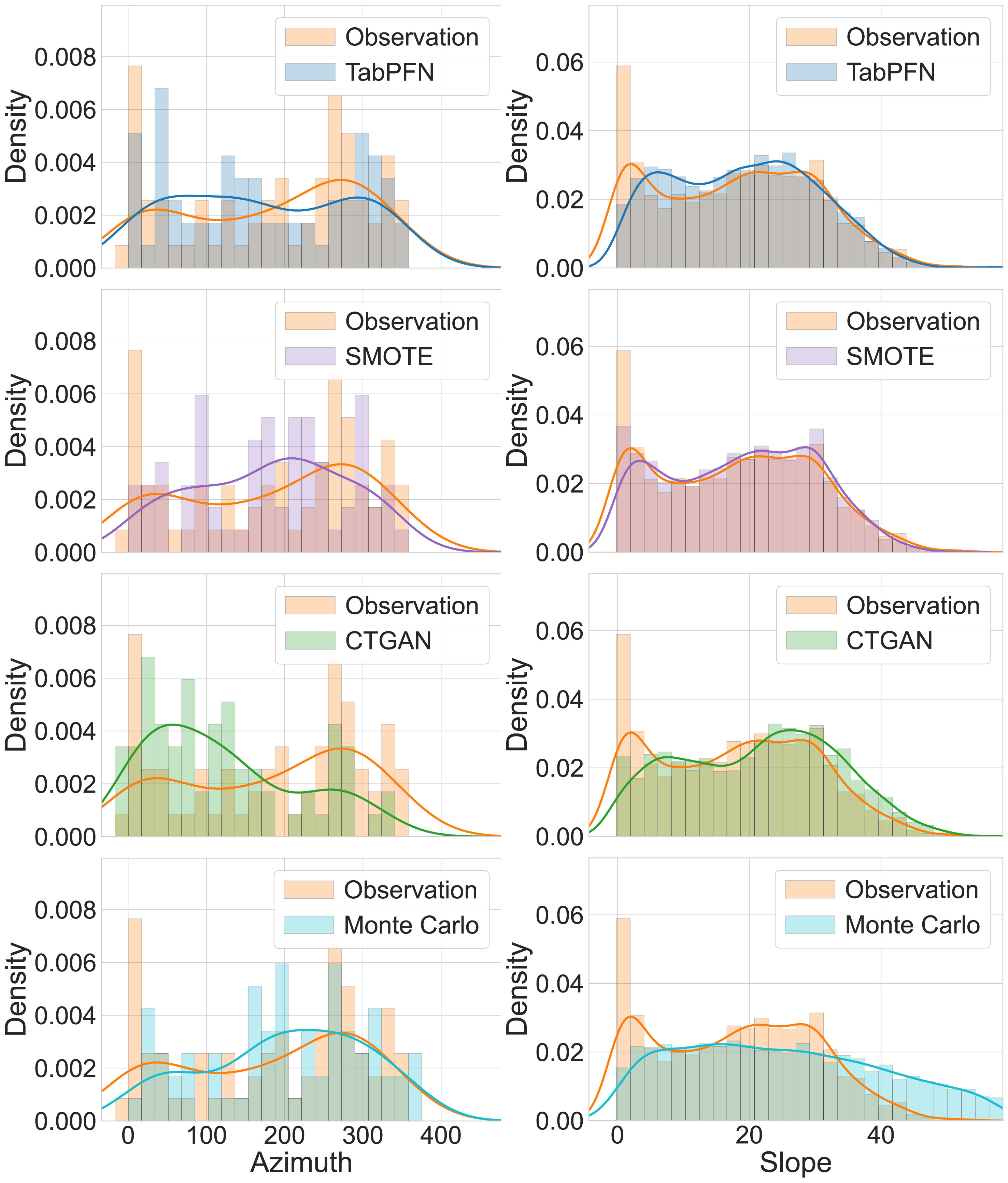}
	\caption{Statistical characteristics of sparse local-scale and abundant global-scale rainfall-triggered landslide inventories}
	\label{fig:4}       
\end{figure}

At global scale, where inventories are large and feature mixtures are more heterogeneous, the slope density reveals a different pattern, TabPFN tends to generate fewer small slope cases than observed, this underrepresentation may reflect that pretrained priors and dominant patterns learned from broad tabular corpora emphasize the more typical steep slope regimes associated with landslides, small slope failures are often controlled by local hydrological or anthropogenic factors that may be underrepresented in the pretraining signal, SMOTE matches small slope density more closely in some instances because it amplifies existing minority neighborhoods in the local data, however advantage of SMOTE can come at the cost of replicating sampling idiosyncrasies and failing to introduce new, physically plausible combinations, these findings imply that combining foundation generation with targeted conditioning or post hoc reweighting of generated samples can correct slope biases while retaining strength of TabPFN in filling structurally missing regions of the joint feature space. 

The proposed approach delivers stable and generalizable performance across varied terrain and geomorphological contexts. It preserves both marginal densities and meaningful multivariate dependencies for numeric and categorical predictors without local pretraining. This robustness is evident in data rich inventories and remains visible under smaller sample sizes and more complex conditional structures. The generated records therefore provide realistic and process consistent augmentations that enhance representativeness and support downstream landslide modeling including susceptibility mapping and conditional probability estimation.

\subsection{Scenario II: Landslide datasets focused on triggering and meteorological characteristics}
\label{sec:3.3}

We also evaluate generation fidelity with respect to triggering and meteorological factors. The analysis focuses on precipitation and humidity related metrics across inventories that range from global datasets with abundant samples to regional datasets with limited observations. Marginal distributions pairwise scatterplots and conditional summaries are examined to assess whether the generated records reproduce meteorological characteristics observed before landslide events and whether they preserve meaningful relationships among these predictors.

\begin{figure}[H]
	\centering
	\includegraphics[width=\textwidth]{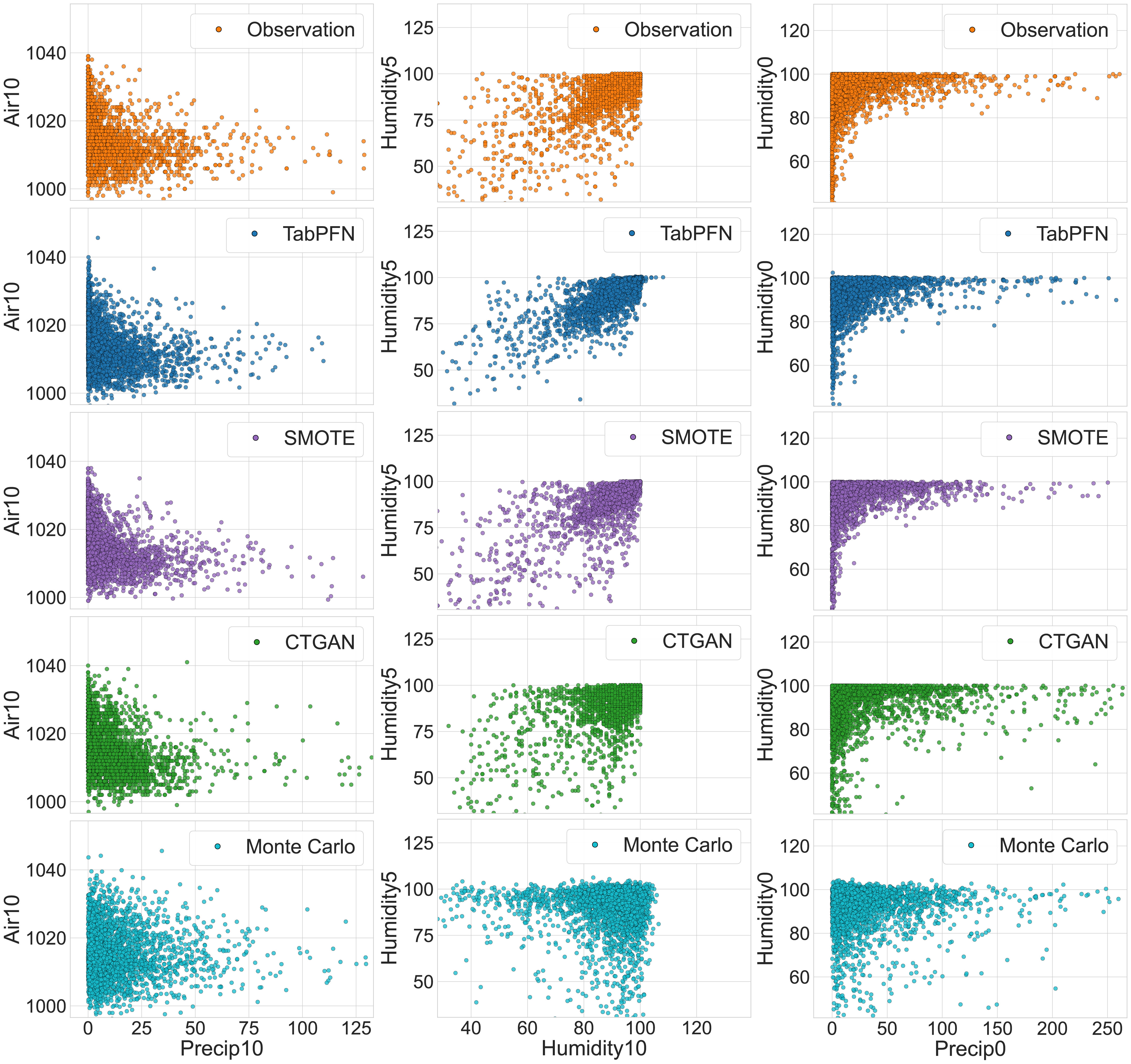}
	\caption{Comparative analysis of meteorological patterns in global-scale rainfall-triggered inventories generated by four approaches}
	\label{fig:5}       
\end{figure}

At the global scale where observations are plentiful short term precipitation variables often display roughly symmetric distributions around climatological norms. Several methods reproduce these simple univariate patterns. However relationships among variables are often more informative than individual predictors. For example humidity measured at two lead times shows a clear positive association in the observed data. Monte Carlo sampling does not capture this association and produces nearly uncorrelated pairs. This behavior generates unrealistic combinations such as intense rainfall together with unusually low humidity. Such inconsistent samples would distort interpretation of moisture related triggering conditions and could bias subsequent conditional analyses. TabPFN and CTGAN generally avoid these contradictions because they preserve the joint structure among meteorological variables. This result suggests that generators which retain multivariate dependencies reduce the risk of producing physically inconsistent events.

More complex joint patterns further highlight differences among methods. Relationships among rainfall accumulations at multiple antecedent windows together with concurrent humidity often show nonlinear coupling. TabPFN preserves these nonlinear dependencies and avoids generating samples that contradict basic hydrometeorological relationships. CTGAN can reproduce some medium and long term rainfall aggregates but is less consistent in capturing short duration maxima that often trigger rapid shallow failures. Monte Carlo sampling lacks the ability to represent such joint complexity because it mainly reproduces marginal distributions. These results indicate that realistic augmentation for rainfall induced landslides requires methods that capture conditional dependence while also representing short term extremes.

In data sparse regional inventories such as the Italy dataset TabPFN also performs consistently across different aggregation windows. The generated samples reproduce relationships between one day rainfall maxima and seven day cumulative rainfall and between thirty day sums and thirty day maxima. This consistency indicates that pretrained multivariate priors together with contextual information can support plausible rainfall signatures even when local observations are limited. CTGAN captures some medium and long term aggregates but tends to underestimate short duration peaks which reduces its usefulness for analyzing rapid triggering processes. Monte Carlo performs poorly across these temporal scales because correct marginal shapes do not guarantee realistic temporal relationships among variables. These findings emphasize that realistic augmentation for sparse inventories requires methods capable of inferring dependencies across multiple temporal windows rather than matching single variable distributions alone.

TabPFN also shows advantages in inventories where landslide occurrences are spatially clustered. In the Chongqing inventory landslide clustering intensity measured using a kernel density metric is associated with both local topography and precipitation conditions. TabPFN reproduces areas of elevated kernel density that correspond to particular combinations of terrain and rainfall features. Other methods often smooth these patterns or interpolate across high density regions which reduces the representation of localized hotspots. Preserving the detailed structure of joint scatterplots and kernel density curves is important because these patterns reflect clustering intensity and tail behavior that influence where failures concentrate. When synthetic data modify these shapes by excessive smoothing or by introducing peaks not present in the observations downstream analyses may misidentify hazard hotspots or misestimate conditional probabilities. TabPFN maintains these localized clustering patterns because pretrained multivariate priors are guided by contextual indicators. This property helps retain spatially coherent signals that are important for translating augmented inventories into reliable landslide susceptibility assessments.

\section{Discussion}
\label{sec:4}

Landslide occurrence reflects the combined effects of terrain conditions, triggering processes, and local geomorphic context, and the available inventories often provide only partial observations of these interactions. As shown in the results, different datasets exhibit distinct marginal shapes and joint feature relationships, and these patterns vary with trigger type, sample size, and regional setting. This makes it necessary to consider not only whether individual variables are reproduced, but also whether the generated data preserve the multivariate structure that carries process related information.

In this study, we evaluate a generative approach that aims to reproduce these observed statistical patterns while extending limited inventories in a controlled way. The analysis focuses on how well the generated samples match observed distributions and relationships across terrain, meteorological, and seismic related features. Based on these results, we further examine the advantages of the method in terms of accuracy, robustness, and transferability, and discuss its limitations and potential directions for future work.

\subsection{Advantages of the proposed generative approach}
\label{sec:4.1}

The proposed generative approach shows three clear advantages in the results, accuracy, robustness, and transferability. Across rainfall triggered and earthquake triggered inventories, the generated records remain close to the observed distributions, preserve meaningful relationships among landslide features, and keep stable performance when the sample size, feature type, and regional setting change. These outcomes indicate that the method can support realistic landslide data augmentation under the sparse and imbalanced conditions that are common in landslide research. The results also suggest that a tabular foundation model is well suited to landslide data, because it can learn from limited observations while still capturing the multivariate structure that ordinary sampling methods often miss.

\subsubsection{Accuracy}
\label{sec:4.1.1}

Accuracy is reflected in the close match between generated records and observed landslide patterns. For both numerical and categorical features, the synthetic data reproduce the main distributional shapes of the original inventories, and they also retain the joint patterns that link different landslide attributes. When the observed data show a clear structure, the generated samples follow that structure well. When the observed distribution is irregular or weakly defined, the proposed method still preserves the main empirical pattern without creating obvious artificial peaks or removing important variation.

Such accuracy is important for landslide data because the scientific meaning of a generated sample depends not only on single variable statistics, but also on whether the sample keeps the same feature relationships as the observation. Landslide inventories often contain complex combinations of terrain, trigger, and geomorphic indicators, and these combinations cannot be represented well by simple interpolation or by sampling each feature separately. The observed accuracy of the proposed method can be attributed to the way the foundation model uses broad pretrained priors to infer likely multivariate structure from limited local data. In this way, the generated records remain close to the real landslide data in both marginal form and joint dependence, which makes them more useful for downstream landslide analysis.

\subsubsection{Robustness}
\label{sec:4.1.2}

Robustness is reflected in the stable behavior of the proposed method across different data sizes and generation settings. The generated records remain consistent even when the original inventory is sparse, uneven, or incomplete. In these cases the method still preserves the main statistical character of the observed data and avoids obvious instability in the generated samples. The results also show that moderate generation multiples improve representativeness, while the generated samples remain close to the original feature space rather than drifting away from it.

The behavior across different generation sizes further supports this robustness. When the number of synthetic samples increases in a controlled way, the augmented inventory becomes more representative and the downstream use of the data becomes more stable. The gain is not from simple repetition, but from adding plausible samples that remain aligned with the observed distribution. This robustness is closely related to the foundation model design, because the model does not need to relearn each inventory from scratch and is less affected by small changes in the local sample set. For landslide research, this is important because sparse inventories often need augmentation, yet the added samples must still preserve the original landslide signal. The proposed method shows that stable augmentation can be achieved without changing the basic statistical character of the original records.

\subsubsection{Transferability}
\label{sec:4.1.3}

Transferability is reflected in the consistent performance of the proposed method across different landslide settings. The same framework works for rainfall triggered and earthquake triggered inventories, even though the two types of events have different dominant controls and different feature structures. In rainfall triggered cases, the generated samples preserve terrain, hydrological, and categorical relationships. In earthquake triggered cases, the generated samples preserve elevation related and geometric patterns. The results show that the method can adapt to different landslide scenarios without changing its basic form.

The transferability also appears in the way the method handles different combinations of feature types. It performs well when numerical and categorical variables appear together, and it remains effective when the dominant relationships vary from one dataset to another. This is important for landslide modeling because no single inventory can represent all regions or all triggers. The strong transferability comes from the tabular foundation model, which is trained to learn general multivariate patterns rather than a single dataset specific rule. Because of this, the method can reuse learned statistical structure across different landslide inventories and can still reflect the local relationships that matter for each case. This ability to generate realistic samples across different settings means the proposed approach can support comparative studies, cross region analysis, and broader landslide modeling tasks while keeping the generated data close to the observed patterns.

\subsection{Limitations and future work}
\label{sec:4.2}

Considering the nature of landslides, the relationship between observed patterns and underlying physical processes is important. The proposed method can reproduce realistic multivariate distributions and preserve feature relationships, yet the generated records are still driven mainly by statistical learning from tabular data rather than by explicit physical rules. As a result, the outputs are useful for representing the observed data structure, but they do not directly explain the geomorphic or mechanical causes behind each sample. Therefore, while the method achieves high consistency with observed data and reflects characteristic patterns and associations of landslide events, its interpretability from a physical and geological perspective remains limited.

In landslide investigation, synthetic records should be viewed as a complement to observations rather than a replacement. The proposed method can support sparse inventories, especially in remote or difficult terrain where complete field surveys are hard to obtain, and can improve the balance and representativeness of training data. At the same time, increasing the number of synthetic samples does not necessarily lead to better scientific understanding. The effectiveness of augmentation still depends on the quality of observed records, the relevance of selected features, and the way generated data are incorporated. In this study, the main objective is to capture intrinsic dependencies among landslide features from limited observations and to use synthetic data as a structured extension of these observations, rather than simply replacing original data with large volumes of generated samples. By extracting key information from observed records and integrating it with generated data, underlying patterns related to landslide occurrence and failure can be more effectively explored.

Future work will focus on making the generated samples more physically informed and more useful for practical analysis. One direction is to integrate the foundation model with stronger geomorphic constraints, expert knowledge, and process based checks, so that generated records remain statistically consistent while becoming more interpretable in landslide terms. Another direction is to develop uncertainty quantification for synthetic data and to examine how different levels of augmentation influence model performance under varying inventory sizes and feature types. These improvements would enhance the transparency and reliability of the approach and support its broader use in addressing data sparsity and imbalance in landslide modeling.

\section{Conclusion}
\label{sec:5}

Sufficient and well-balanced landslide datasets are essential for understanding landslide occurrence and failure processes. In this study, we propose a simple and robust approach for generating multi feature landslide datasets to overcome data sparsity and imbalance. The key idea is to capture complex distributional patterns and inter-feature relationships from limited observations based on foundation model, and to generate distinctive and informative samples that help overcome landslide data sparsity and imbalance. Based on comparative experiments on 20 rainfall and earthquake triggered landslide inventories, we found that: 

(1) The proposed approach captures multi-feature dependencies from limited observations and thereby reproduces key statistical characteristics.

(2) The proposed approach performs well across rainfall and earthquake triggered landslides from different regions demonstrating robustness and transferability. 

(3) The proposed approach captures latent associations from sparse samples and generates distinctive landslide cases that help overcome data sparsity and imbalance.

\section*{Declaration of competing interest}

The authors declare that they have no known competing financial interests or personal relationships that could have appeared to influence the work reported in this paper.

\section*{Acknowledgments}

This work was supported by the National Natural Science Foundation of China (Grant No.42277161).

\newpage


\section*{References}

\bibliographystyle{elsarticle-harv}           
\bibliography{reference}

\begin{thebibliography}{43}
\expandafter\ifx\csname natexlab\endcsname\relax\def\natexlab#1{#1}\fi
\expandafter\ifx\csname url\endcsname\relax
  \def\url#1{\texttt{#1}}\fi
\expandafter\ifx\csname urlprefix\endcsname\relax\def\urlprefix{URL }\fi

\bibitem[{Ado et~al.(2022)Ado, Amitab, Maji, et~al.}]{Ado2022}
Ado, M., Amitab, K., Maji, A., et~al., 2022. Landslide susceptibility mapping
  using machine learning: a literature survey. Remote Sensing 14, 3029.

\bibitem[{Al-Najjar et~al.(2021)Al-Najjar, Pradhan, Sarkar,
  et~al.}]{AlNajjar2021}
Al-Najjar, H., Pradhan, B., Sarkar, et~al., 2021. A new integrated approach for
  landslide data balancing and spatial prediction based on generative
  adversarial networks (gan). Remote Sensing 13, 4011.

\bibitem[{Alkhasawneh and Tay(2018)}]{Alkhasawneh2018}
Alkhasawneh, M., Tay, L., 2018. A hybrid intelligent system integrating the
  cascade forward neural network with elman neural network. Arab Journal of
  Science and Engineering 43, 6737--6749.

\bibitem[{Althuwaynee et~al.(2014)Althuwaynee, Pradhan, Park, and
  Lee}]{Althuwaynee2014}
Althuwaynee, O., Pradhan, B., Park, H., Lee, J., 2014. A novel ensemble
  decision tree-based chi-squared automatic interaction detection (chaid) and
  multivariate logistic regression models in landslide susceptibility mapping.
  Landslides 11, 1063--1078.

\bibitem[{Azarafza et~al.(2021)Azarafza, Azarafza, Akgün,
  et~al.}]{Azarafza2021}
Azarafza, M., Azarafza, M., Akgün, H., et~al., 2021. Deep learning-based
  landslide susceptibility mapping. Scientific Reports 11, 24112.

\bibitem[{Bodnar et~al.(2025)Bodnar, Bruinsma, Lucic, Stanley, Allen,
  Brandstetter, Garvan, Riechert, Weyn, Dong, Gupta, Thambiratnam, Archibald,
  Wu, Heider, Welling, Turner, and Perdikaris}]{Bodnar2025}
Bodnar, C., Bruinsma, W.~P., Lucic, A., Stanley, M., Allen, A., Brandstetter,
  J., Garvan, P., Riechert, M., Weyn, J.~A., Dong, H., Gupta, J.~K.,
  Thambiratnam, K., Archibald, A.~T., Wu, C.-C., Heider, E., Welling, M.,
  Turner, R.~E., Perdikaris, P., 2025. A foundation model for the earth system.
  Nature 641~(8065), 1180–1187.

\bibitem[{Breiman(1996)}]{Breiman1996}
Breiman, L., 1996. Bagging predictors. Machine Learning 24, 123--140.

\bibitem[{Chawla(2010)}]{Chawla2010}
Chawla, N.~V., 2010. Data mining for imbalanced datasets: an overview. In: Data
  Mining and Knowledge Discovery Handbook. Springer US.

\bibitem[{Conoscenti et~al.(2016)Conoscenti, Rotigliano, Cama, Caraballo-Arias,
  Lombardo, and Agnesi}]{Conoscenti2016}
Conoscenti, C., Rotigliano, E., Cama, M., Caraballo-Arias, N., Lombardo, L.,
  Agnesi, V., 2016. Exploring the effect of absence selection on landslide
  susceptibility models: a case study in sicily, italy. Geomorphology 261,
  222--235.

\bibitem[{Du et~al.(2020)Du, Glade, Woldai, Chai, and Zeng}]{Du2020}
Du, J., Glade, T., Woldai, T., Chai, B., Zeng, B., 2020. Landslide
  susceptibility assessment based on an incomplete landslide inventory in the
  jilong valley, tibet, chinese himalayas. Engineering Geology 270, 105572.

\bibitem[{Fang et~al.(2021)Fang, Wang, Niu, and Peng}]{Fang2021}
Fang, Z., Wang, Y., Niu, R., Peng, L., 2021. Landslide susceptibility
  prediction based on positive unlabeled learning coupled with adaptive
  sampling. IEEE Journal of Selected Topics in Applied Earth Observations and
  Remote Sensing 14, 11581--11592.

\bibitem[{Freund and Schapire(1995)}]{Freund1995}
Freund, Y., Schapire, R.~E., 1995. A decision-theoretic generalization of
  on-line learning and an application to boosting. In: Computational Learning
  Theory. Springer.

\bibitem[{Galar et~al.(2012)Galar, Fernandez, Barrenechea, et~al.}]{Galar2012}
Galar, M., Fernandez, A., Barrenechea, E., et~al., 2012. A review on ensembles
  for the class imbalance problem: bagging-, boosting-, and hybrid-based
  approaches. IEEE Transactions on Systems, Man, and Cybernetics, Part C
  (Applications and Reviews) 42, 463--484.

\bibitem[{Glade et~al.(2012)Glade, Anderson, and Crozier}]{Glade2012}
Glade, T., Anderson, M., Crozier, M.~J., 2012. Landslide Hazard and Risk. John
  Wiley $\&$ Sons Ltd.

\bibitem[{Goetz et~al.(2015)Goetz, Brenning, Petschko, and Leopold}]{Goetz2015}
Goetz, J., Brenning, A., Petschko, H., Leopold, P., 2015. Evaluating machine
  learning and statistical prediction techniques for landslide susceptibility
  modeling. Computers $\&$ Geosciences 81, 1--11.

\bibitem[{Gupta et~al.(2022)Gupta, Pal, and Das}]{Gupta2022}
Gupta, N., Pal, S., Das, J., 2022. Gis-based evolution and comparisons of
  landslide susceptibility mapping of the east sikkim himalaya. Annals of GIS
  28~(3), 359--384.

\bibitem[{Gupta et~al.(2020)Gupta, Jhunjhunwalla, Bhardwaj, and
  Shukla}]{Gupta2020}
Gupta, S., Jhunjhunwalla, M., Bhardwaj, A., Shukla, D., 2020. Data imbalance in
  landslide susceptibility zonation: under-sampling for class-imbalance
  learning. In: ISPRS - International Archives of the Photogrammetry, Remote
  Sensing and Spatial Information Sciences. Vol. XLII-3/W11. pp. 51--57.

\bibitem[{Guzzetti et~al.(2012)Guzzetti, Mondini, Cardinali, Fiorucci,
  Santangelo, and Chang}]{Guzzetti2012}
Guzzetti, F., Mondini, A.~C., Cardinali, M., Fiorucci, F., Santangelo, M.,
  Chang, K.-T., 2012. Landslide inventory maps: new tools for an old problem.
  Earth-Science Reviews 112, 42--66.

\bibitem[{Haixiang et~al.(2017)Haixiang, Yijing, Shang, et~al.}]{Haixiang2017}
Haixiang, G., Yijing, L., Shang, J., et~al., 2017. Learning from
  class-imbalanced data: review of methods and applications. Expert Systems
  with Applications 73, 220--239.

\bibitem[{He and Garcia(2009)}]{He2009}
He, H., Garcia, E.~A., 2009. Learning from imbalanced data. IEEE Transactions
  on Knowledge and Data Engineering 21, 1263--1284.

\bibitem[{Hollmann et~al.(2025)Hollmann, Müller, Purucker, Krishnakumar,
  Körfer, Hoo, Schirrmeister, and Hutter}]{Hollmann2025}
Hollmann, N., Müller, S., Purucker, L., Krishnakumar, A., Körfer, M., Hoo,
  S.~B., Schirrmeister, R.~T., Hutter, F., 2025. Accurate predictions on small
  data with a tabular foundation model. Nature 637~(8045), 319–326.

\bibitem[{Hong et~al.(2007)Hong, Adler, and Huffman}]{Hong2007}
Hong, Y., Adler, R., Huffman, G., 2007. Satellite remote sensing for global
  landslide monitoring. Eos (Washington DC) 88, 357--358.

\bibitem[{Huang et~al.(2020)Huang, Luo, and Lin}]{Huang2020}
Huang, L., Luo, J., Lin, Z. e.~a., 2020. Using deep learning to map
  retrogressive thaw slumps in the beiluhe region (tibetan plateau) from
  cubesat images. Remote Sensing of Environment 237, 111534.

\bibitem[{Hussin et~al.(2016)Hussin, Zumpano, Reichenbach, Sterlacchini, Micu,
  van Westen, and Bălteanu}]{Hussin2016}
Hussin, H., Zumpano, V., Reichenbach, P., Sterlacchini, S., Micu, M., van
  Westen, C., Bălteanu, D., 2016. Different landslide sampling strategies in a
  grid-based bi-variate statistical susceptibility model. Geomorphology 253,
  508--523.

\bibitem[{Lee et~al.(2018)Lee, Sameen, Pradhan, and Park}]{Lee2018}
Lee, J., Sameen, M., Pradhan, B., Park, H., 2018. Modeling landslide
  susceptibility in data-scarce environments using optimized data mining and
  statistical methods. Geomorphology 303, 284--298.

\bibitem[{Liu et~al.(2009)Liu, Wu, and Zhou}]{Liu2009}
Liu, X.-Y., Wu, J., Zhou, Z.-H., 2009. Exploratory undersampling for
  class-imbalance learning. IEEE Transactions on Systems, Man, and Cybernetics,
  Part B 39, 539--550.

\bibitem[{Ma et~al.(2021)Ma, Mei, and Piccialli}]{Ma2021}
Ma, Z., Mei, G., Piccialli, F., 2021. Machine learning for landslides
  prevention: a survey. Neural Computing and Applications 33, 10881--10907.

\bibitem[{Micheletti et~al.(2014)Micheletti, Foresti, and
  Robert}]{Micheletti2014}
Micheletti, N., Foresti, L., Robert, S. e.~a., 2014. Machine learning feature
  selection methods for landslide susceptibility mapping. Mathematical
  Geosciences 46, 33--57.

\bibitem[{Nanni et~al.(2015)Nanni, Fantozzi, and Lazzarini}]{Nanni2015}
Nanni, L., Fantozzi, C., Lazzarini, N., 2015. Coupling different methods for
  overcoming the class imbalance problem. Neurocomputing 158, 48--61.

\bibitem[{Nikoobakht et~al.(2022)Nikoobakht, Azarafza, Akgün, and
  Derakhshani}]{Nikoobakht2022}
Nikoobakht, S., Azarafza, M., Akgün, H., Derakhshani, R., 2022. Landslide
  susceptibility assessment by using convolutional neural network. Applied
  Sciences 12, 5992.

\bibitem[{Petschko et~al.(2014)Petschko, Brenning, and Bell}]{Petschko2014}
Petschko, H., Brenning, A., Bell, R. e.~a., 2014. Assessing the quality of
  landslide susceptibility maps - case study lower austria. Natural Hazards and
  Earth System Sciences 14, 95--118.

\bibitem[{Polikar(2012)}]{Polikar2012}
Polikar, R., 2012. Ensemble learning. In: Ensemble Machine Learning. Springer,
  pp. 1--34.

\bibitem[{Sameen et~al.(2020)Sameen, Pradhan, Bui, and Alamri}]{Sameen2020}
Sameen, M., Pradhan, B., Bui, D., Alamri, A., 2020. Systematic sample
  subdividing strategy for training landslide susceptibility models. Catena
  187, 104358.

\bibitem[{Song et~al.(2018)Song, Niu, Xu, et~al.}]{Song2018}
Song, Y., Niu, R., Xu, S., et~al., 2018. Landslide susceptibility mapping based
  on weighted gradient boosting decision tree in wanzhou section of the three
  gorges reservoir area (china). ISPRS International Journal of Geo-Information
  8, 4.

\bibitem[{Steger et~al.(2016)Steger, Brenning, Bell, and Glade}]{Steger2016}
Steger, S., Brenning, A., Bell, R., Glade, T., 2016. The influence of
  systematically incomplete shallow landslide inventories on statistical
  susceptibility models and suggestions for improvements. Landslides 14,
  1767--1781.

\bibitem[{Tang et~al.(2009)Tang, Zhang, and Chawla}]{Tang2009}
Tang, Y., Zhang, Y., Chawla, N., 2009. Svms modeling for highly imbalanced
  classification. IEEE Transactions on Systems, Man, and Cybernetics, Part B:
  Cybernetics 39, 281--288.

\bibitem[{Taylor et~al.(2018)Taylor, Malamud, Witt, and Guzzetti}]{Taylor2018}
Taylor, F.~E., Malamud, B.~D., Witt, A., Guzzetti, F., 2018. Landslide shape,
  ellipticity and length-to-width ratios. Earth Surface Processes and Landforms
  43, 3164--3189.

\bibitem[{Wang et~al.(2019)Wang, Wu, Chen, et~al.}]{Wang2019}
Wang, Y., Wu, X., Chen, Z., et~al., 2019. Optimizing the predictive ability of
  machine learning methods for landslide susceptibility mapping using smote for
  lishui city in zhejiang province, china. International Journal of
  Environmental Research and Public Health 16, 368.

\bibitem[{Yao et~al.(2022)Yao, Qin, Qiao, et~al.}]{Yao2022}
Yao, J., Qin, S., Qiao, S., et~al., 2022. Application of a two-step sampling
  strategy based on deep neural network for landslide susceptibility mapping.
  Bulletin of Engineering Geology and the Environment 81, 148.

\bibitem[{Zhong et~al.(2020)Zhong, Liu, and Gao}]{Zhong2020}
Zhong, C., Liu, Y., Gao, P. e.~a., 2020. Landslide mapping with remote sensing:
  challenges and opportunities. International Journal of Remote Sensing 41,
  1555--1581.

\bibitem[{Zhu et~al.(2019)Zhu, Miao, Liu, Bai, Zeng, Ma, and Hong}]{Zhu2019}
Zhu, A., Miao, Y., Liu, J., Bai, S., Zeng, C., Ma, T., Hong, H., 2019. A
  similarity-based approach to sampling absence data for landslide
  susceptibility mapping using data-driven methods. Catena 183, 104188.

\bibitem[{Zhu et~al.(2026{\natexlab{a}})Zhu, Zhang, Wei, Wang, Liu, Yan, and
  Wang}]{Zhu20261}
Zhu, H., Zhang, X., Wei, G., Wang, Q., Liu, X., Yan, L., Wang, G.,
  2026{\natexlab{a}}. A method for better mapping of susceptibility to thaw
  hazards in data-scarce cold regions. Remote Sensing of Environment 337,
  115338.

\bibitem[{Zhu et~al.(2026{\natexlab{b}})Zhu, Xiong, Wang, Stewart, Heidler,
  Wang, Yuan, Dujardin, Xu, and Shi}]{Zhu20262}
Zhu, X.~X., Xiong, Z., Wang, Y., Stewart, A.~J., Heidler, K., Wang, Y., Yuan,
  Z., Dujardin, T., Xu, Q., Shi, Y., 2026{\natexlab{b}}. On the foundations of
  earth foundation models. Communications Earth $\&$ Environment 7~(1), 103.

\end{thebibliography}




\end{document}